\documentclass[10pt,twocolumn,letterpaper]{article}

\usepackage{cvpr}              % To produce the CAMERA-READY version
\usepackage{graphicx}
\usepackage{amsmath}
\usepackage{amssymb}
\usepackage{booktabs}
\usepackage{url}            % simple URL typesetting
\usepackage{amsfonts}       % blackboard math symbols
\usepackage{nicefrac}       % compact symbols for 1/2, etc.
\usepackage{microtype}      % microtypography
\usepackage{xcolor}         % colors
\usepackage{algorithm,algpseudocode}
\usepackage{wrapfig}
\usepackage{multirow}
\usepackage{float}

\newcommand{\cut}[1]{}

\newtheorem{problem}{Problem}

\newcommand\blfootnote[1]{
    \begingroup
    \renewcommand\thefootnote{}\footnote{#1}
    \addtocounter{footnote}{-1}
    \endgroup
}
\usepackage[pagebackref,breaklinks,colorlinks]{hyperref}

\usepackage[capitalize]{cleveref}
\crefname{section}{Sec.}{Secs.}
\Crefname{section}{Section}{Sections}
\Crefname{table}{Table}{Tables}
\crefname{table}{Tab.}{Tabs.}

\def\cvprPaperID{9721} % *** Enter the CVPR Paper ID here
\def\confName{CVPR}
\def\confYear{2022}

\begin{document}

\title{Meta Adaptive Task Sampling for Few-Domain Generalization}

\author{Zheyan Shen \textsuperscript{1}$^{\dag}$,
Han Yu \textsuperscript{1}$^{\dag}$,
Peng Cui \textsuperscript{1}$^*$,
Jiashuo Liu \textsuperscript{1},
Xingxuan Zhang \textsuperscript{1},
Linjun Zhou \textsuperscript{1},
Furui Liu \textsuperscript{2}\\
\textsuperscript{1}Department of Computer Science and Technology, Tsinghua University\\
\textsuperscript{2}Huawei Noah’s Ark Lab\\
{\tt\small shenzy13@qq.com, yuh21@mails.tsinghua.edu.cn, cuip@tsinghua.edu.cn, liujiashuo77@gmail.com}\\ {\tt\small xingxuanzhang@hotmail.com, zhoulj16@mails.tsinghua.edu.cn, 
liufurui2@huawei.com}
}
\maketitle

%%%%%%%%% ABSTRACT
\begin{abstract}
To ensure the out-of-distribution (OOD) generalization performance, traditional domain generalization (DG) methods resort to training on data from multiple sources with different underlying distributions.
And the success of those DG methods largely depends on the fact that there are diverse training distributions.
However, it usually needs great efforts to obtain enough heterogeneous data due to the high expenses, privacy issues or the scarcity of data.
Thus an interesting yet seldom investigated problem arises: how to improve the OOD generalization performance when the perceived heterogeneity is limited.
In this paper, we instantiate a new framework called few-domain generalization (FDG), which aims to learn a generalizable model from very few domains of novel tasks with the knowledge acquired from previous learning experiences on base tasks.
Moreover, we propose a Meta Adaptive Task Sampling (MATS) procedure to differentiate base tasks according to their semantic and domain-shift similarity to the novel task.
Empirically, we show that the newly introduced FDG framework can substantially improve the OOD generalization performance on the novel task and further combining MATS with episodic training could outperform several state-of-the-art DG baselines on widely used benchmarks like PACS and DomainNet.
\end{abstract}
\blfootnote{$\dag$Equal contribution, *Corresponding author}

\section{Introduction}
\label{sec:intro}
%第一段：简述独立同分布的局限性，引出OOD泛化的问题及重要性
The promising results of most machine learning methods actually rely on a bedrock that the data encountered at testing phase are drawn from the same distribution as those the model trained on (a.k.a. I.I.D. hypothesis).
However, once we can not fully control the data generation process, which is inevitable in many real scenarios, distribution shift (or domain shift) problem would naturally arise between the source data on which you train your model and the target data on which the model is deployed.
As a consequence, the model trained only on source data will deteriorate drastically in terms of test performance on target data \cite{torralba2011unbiased}, triggering a crucial problem called out-of-distribution (OOD) generalization. 

%第二段：简要回顾OOD问题的两种常见研究范式DA和DG
On tackling such issue, domain adaptation techniques have been intensively developed during the last two decades by assuming the access to instances (whether labeled or not) from the target distribution on which we deploy our model \cite{shimodaira2000improving, bickel2009discriminative, fernando2013unsupervised, tzeng2017adversarial}.
Typically, a transformation is learned to align the source and target distribution by some kind of distance metric.
Despite the theoretical guarantee of performance produced by these learning methods \cite{ben2010theory}, domain adaptation model cannot generalize to unseen domains by nature, which limits its applications.
Therefore a more challenging problem, domain generalization \cite{blanchard2011generalizing, muandet2013domain, Wang2021, Zhou2021}, has become the focus of research attention where the target (test) distribution is unknown.
By utilizing the training data collected from multiple domains, domain generalization methods can learn an orthogonal decomposition of model parameters \cite{khosla2012undoing, li2017deeper} or, more directly, learn an invariant representation across different domains \cite{smola2007hilbert, motiian2017unified, li2018domain}.

%第三段：现有DG研究针对Domain数量K比较多的场景有不错的表现，但对于K比较少的场景效果无法保证
While a myriad of algorithms have been developed in domain generalization, their empirical performance often largely relies on the sufficiency of data heterogeneity (i.e. domain labels), as illustrated in \cite{2021How} that it is important to have diverse training distributions for out-of-distribution (OOD) generalization.
The learning-theoretic bound developed under kernel space \cite{muandet2013domain} also indicates the positive influence by increasing the visible domains as it can control the OOD generalization error.
Nevertheless, it might not be quite easy to obtain enough domain labels due to the high expenses, privacy issues or just because the data are scarce and hard to collect.
Therefore, an important yet seldom investigated problem is how to improve the OOD generalization performance when the perceived heterogeneity cannot fully support the conventional domain generalization algorithms.
\begin{figure*}
    \centering
	\includegraphics[width=0.8\linewidth]{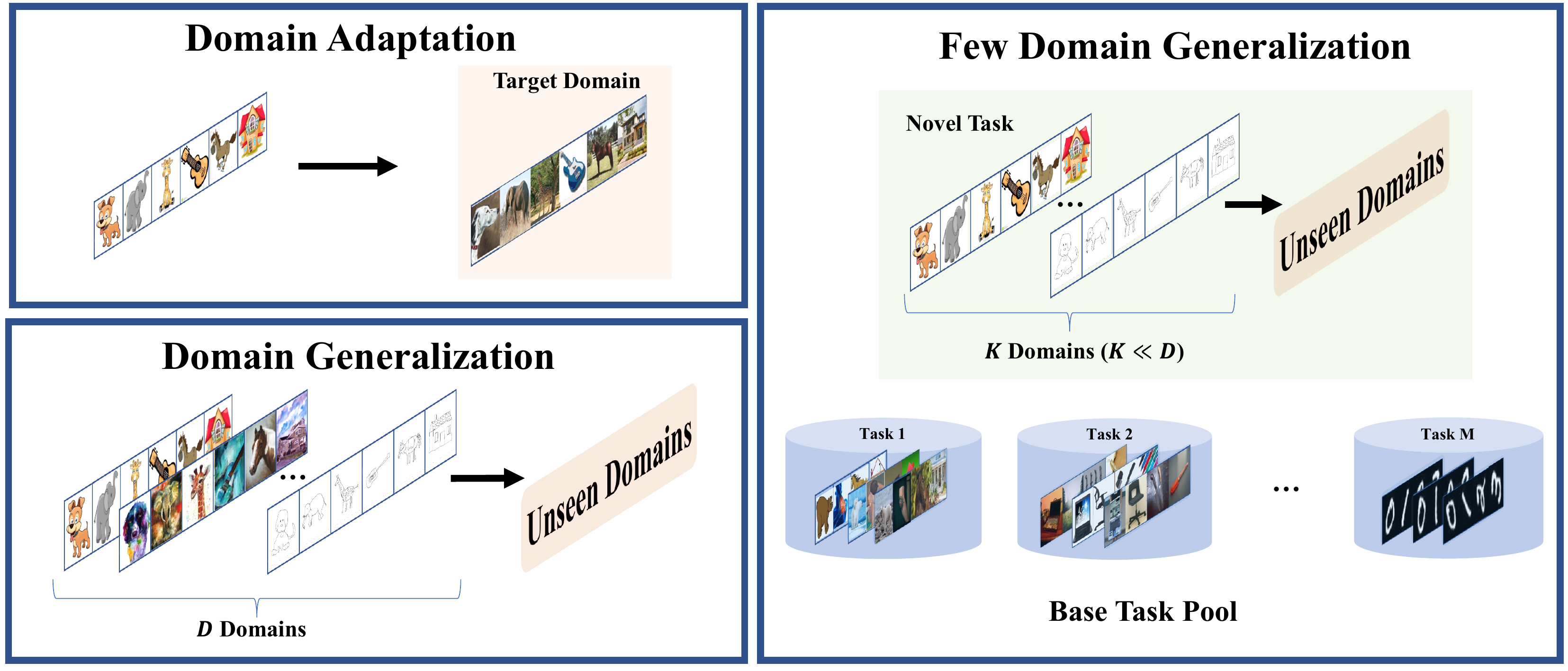}
	\caption{Comparisons of different learning paradigms.}
	\label{fig:problem}
\end{figure*}

%第四段：用实际场景中人类学习的例子引出FDG的问题
Very recently, there exist few methods aiming at generalizing from a single domain \cite{matsuura2020domain, qiao2020learning}, which is an extreme case that there is no explicit heterogeneity at all.
And a straight-forward way to accomplish such goal is to augment the original sample and generate fictitious domains.
Here, we investigate this problem through the lens of humans' learning behavior.
For example, when humans learn how to recognize animals, being exposed to different environments (e.g. light condition, backgound or wheather) does not only improve the generalization of animal recognition tasks.
More crucially, such learning experiences also enable humans to quickly adapt to other tasks (e.g. vehicle recognition) with improved generalization ability, even when the data and heterogeneity from novel task are limited.
Motivated by such intuition, we propose to instantiate a new generalization framework, few-domain generalization, which aims to learn a generalizable model from very few domains with the external experiences on previous tasks.
The differences between few-domain generalization and other common frameworks are illustrated in Figure \ref{fig:problem}.

%第五段：总结本文的贡献：提出了一个新的问题，更具实际意义；提出了一种算法；实验上验证了问题本身不是琐碎的，并证明了算法的有效性
We first empirically confirm that the newly introduced FDG framwork which leverages the base tasks can substantially improve the OOD generalization performance on novel tasks.
However, the conventional algorithms treat all the base tasks equally, which cause the inefficiency at pre-training stage.
Here we argue that the base tasks do not equally contribute to a given novel task and should be differentiated during the pre-training phase.
To address this issue, we propose a Meta Adaptive Task Sampling (MATS) procedure based on episodic training to sample training episodes according to their semantic and domain-shift similarity to the novel task.
As a result, the base tasks which share closer semantic space or more similar domain shift pattern would be up-weighted during meta pre-training.
Empirically, we show that MATS can outperform several state-of-the-art DG baselines on few-domain generalization settings constructed from widely used benchmarks like PACS and DomainNet.
 
%\begin{table*}[t]
%	\small
%	\caption{Taxonomy for domain adaptation (DA), domain generalization (DG) and few-domain generalization (FDG).}
%	\begin{center}
%		\[\arraycolsep=0.8pt\def\arraystretch{1.5}
%		\begin{tabular}{c||c|c}
%			problem & training data  & test data \\\hline 
%			Domain Adaptation (DA)& $(\B{X}^S, {Y}^S),\B{X}^T$ & $(\B{X}^T, {Y}^T)$\\\hline
%			Domain Generalization (DG)& $(\B{X}^1, {Y}^1), \ldots, (\B{X}^M, {Y}^M)$ & $(\B{X}^{M+1}, {Y}^{M+1}), (\B{X}^{M+2}, {Y}^{M+2}), \ldots$\\\hline
%			\multirow{2}{*}{Few-Domain Generalization (FDG)}& $(\B{X}^1, {Y}^1), \ldots, (\B{X}^M, {Y}^M)\in Task1 $& \multirow{2}{*}{$(\B{X}^{D+1}, {Y}^{D+1}), (\B{X}^{D+2}, {Y}^{D+2}), \ldots \in Task2$}\\
%			& $(\B{X}^1, {Y}^1), \ldots, (\B{X}^D, {Y}^D) \in Task2, D \ll M  $& 
%		\end{tabular}
%		\]
%	\end{center}
%	\label{tab:compare}
%\end{table*}
\section{Related Work}

\begin{figure*}[t]
    \centering	
	\includegraphics[width=0.9\linewidth]{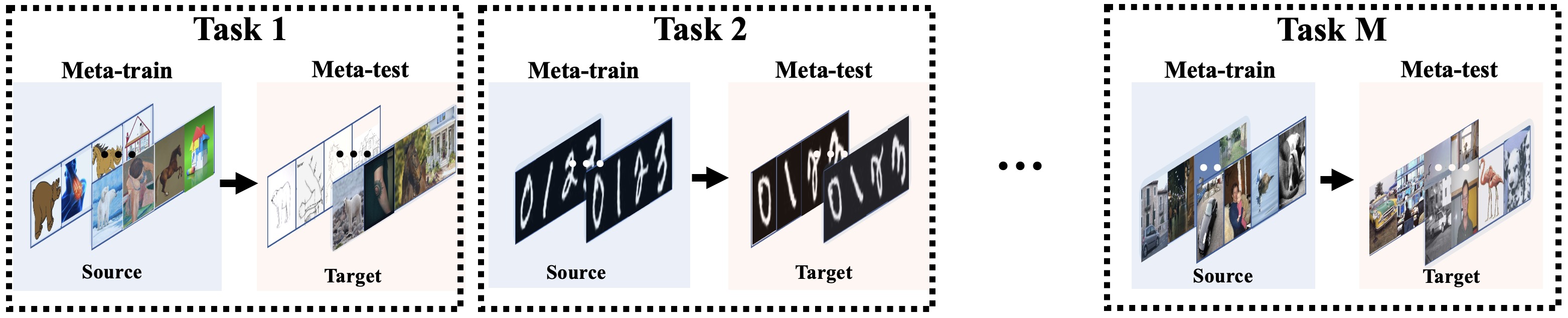}
	\caption{Episodic training on base tasks.}
	\label{fig:mldg_fdg}
\end{figure*}

\label{sec:relate}
In this section, we investigate and compare several related topics more thoroughly, including domain adaptation (DA), domain generalization (DG) and meta-learning.

\textbf{Domain Adaptation (DA)} is one of the most straightforward ways to improve the performance on new target domains provided that you have the prior knowledge on those domains.
It has received great attention from different communities like machine learning, data mining, computer vision, etc.
The key concept of domain adaptation is to align the data or model between source domain and target domain.
Approaches in early stages mainly focus on reweighting samples to match the data distribution \cite{shimodaira2000improving} between domains, leveraging different density ratio estimation methods \cite{bickel2009discriminative, dudik2006correcting, huang2007correcting}.
More recently, with the advances of representation learning techniques (e.g. deep learning), more and more methods try to narrow the discrepancy between source and target domains in the embedding space \cite{fernando2013unsupervised, ganin2014unsupervised, kumar2010co, ganin2016domain, long2016unsupervised, tzeng2017adversarial}, either using maximum mean discrepancy \cite{gretton2012optimal} or adversarial training \cite{goodfellow2014generative}.
Despite the theoretical guarantee of these learning methods \cite{ben2010theory} on target domain performance, domain adaptation model cannot generalize to unseen domains, which limits its applications in most of online scenarios.

\textbf{Domain Generalization (DG)} \cite{Wang2021, Zhou2021} closely relates to domain adaptation in that they both care about the performance of target domains rather than source domains.
However, in domain generalization, we do not assume the availability of labeled (or unlabeled) samples from the target domain, which allows the target domain to be unseen and agnostic.
Most existing DG approches can be divided into three categories.
The first strand of methods rely on a basic assumption that a domain can be decomposed into two parts: domain-agnostic component and domain-specific component.
By learning an orthogonal decomposition on the training source domains \cite{khosla2012undoing, li2017deeper}, the domain-agnostic parameters can therefore be applied to unseen target domains with minimal domain bias.
The second line of research focuses on finding a domain-invariant representation that can minimize the distribution discrepancy between multiple source domains under some types of distance space.
For example, Muandet et al. \cite{muandet2013domain} propose a kernel-based optimization algorithm to learn an invariant representation in reproducing kernel Hilbert space (RKHS) \cite{smola2007hilbert}.
Ghifary et al. \cite{ghifary2015domain} leverage multi-head auto-encoder to learn a general representation that can well reconstruct sample pairs from different domains.
Other techniques like contrastive loss \cite{motiian2017unified} and adversarial anto-encoder \cite{li2018domain} can also be exploited for the same purpose.
The third way to improve generalization ability is to exploit data augmentation in the training phase.
Shankar et al. \cite{shankar2018generalizing} incorporate Bayesian network to generate perturbated samples in a gradient-based scheme.
Volpi et al. \cite{volpi2018generalizing} propose a distributionally robust optimization (DRO) framework to generate adversarial samples.
More recently, there exist a few methods aiming at generalizing from a single domain \cite{matsuura2020domain, qiao2020learning}, which are very similar to our target problem.
Nevertheless, we argue that the heterogeneity of data distribution generated from only one domain can hardly be guaranteed, resulting in inconsistent performance over different datasets.

\textbf{Meta-learning} or learning to learn has a long history which can be traced back to last centry when researchers were interested in training a meta-learner that could train models itself \cite{schmidhuber1987evolutionary, schmidhuber1997shifting, Thrun:1998:LL:296635}.
Recently, meta-learning has attracted a lot of attention due to its good performance on several applications such as parameter generation \cite{li2016learning}, optimizer transfer \cite{andrychowicz2016learning, ravi2016optimization} and few-shot learning \cite{finn2017model, vinyals2016matching, nichol2018first, snell2017prototypical}.
Among these methods, the model-agnostic meta-learning (MAML) \cite{finn2017model} which introduces the concept of ``episodes" in the training phase has greatly influenced the research of domain generalization.
By leveraging the episodic training strategy, several meta-learning methods have been proposed to address the generalization performance on unseen domains \cite{li2018learning, li2019episodic, balaji2018metareg, li2019feature, dou2019domain}.
As noticed in \cite{balaji2018metareg}, such MAML-like training stratey is designed for fast task adaptation using the meta-learned weight initialization (e.g. as in few-shot learning).
Yet traditional domain generalization actually acts as a zero-shot learning problem in that we do not have data from target domains.
In contrast, our few-domain generalization problem may fit the episodic training strategy better and we could anticipate that model can generalize to unseen domains after a small number of gradient descent steps on new task.

% \textbf{Multi-Task Learning (MTL)} tries to learn multiple related tasks simultaneously by extracting shared knowledge (e.g. common feature representations) that can benefit an individual task \cite{caruana1997multitask, argyriou2008convex}.
% However, multi-task learning assumes the training and test data are independent and identically distributed (I.I.D.), which is violated in domain generalization scenario.
% Moreover, it treats all the tasks equally.
% In few-domain generalization, we care more about the novel tasks where the heterogeneity is insufficient, rather than the base tasks.

\section{Problem and Methodology}
\label{sec:method}
In this section, we will first introduce the formal definition of few-domain generalization problem.
Subsequently, we will present a direct application of episodic training strategy on FDG.
Finally we will propose our Meta Adaptive Task Sampling (MATS) procedure to efficiently learn a robust representation which can be quickly adapted to novel tasks with good generalization ability.

\subsection{Problem Setup}
\label{subsec:setup}
\textbf{Notation}: Let $\mathcal{X}$ be the feature space and $\mathcal{Y}$ the label space, and a \emph{domain} is defined as a joint distribution $P_{XY}$ on $\mathcal{X} \times \mathcal{Y}$.
A parametric model is defined as $f: \mathcal{X} \to \mathcal{Y}$, which could be further divided into two parts: a feature extractor $\Theta_{\theta}(\cdot)$ and a classifier $\Psi_{\psi}(\cdot)$, so that $f(\mathbf{x})=\Psi_{\psi}(\Theta_{\theta}(\mathbf{x}))$.
We also have a task space $\mathcal{T}$ where we can sample tasks from it.
Each task $T \in \mathcal{T}$ is a \emph{multi-domain} dataset consisting of $K_T$ domains {$\mathcal{D}^T = \{D_k^T\footnote{For simplicity, we will omit task indicator $T$ for each given domain in the following.}\}_{k=1}^{K_T}$}, which can be further divided into source and target domains as in domain generalization settings, i.e. $\mathcal{D}^T$ = $\mathcal{D}^T_{src} \cup \mathcal{D}^T_{tar}$.
Each domain $k$ in the task contains $N_k$ i.i.d. data points sampled from underlying joint distribution $P_{XY}^{(k)}$, namely $D_k = \{(x_i^{(k)}, y_i^{(k)})\}_{i=1}^{N_k}$ with $(x_i^{(k)}, y_i^{(k)}) \sim P_{XY}^{(k)}$.
In general, $P_{XY}^{(k)} \neq P_{XY}^{(k^\prime)}$ with $k \neq k^\prime$ and $k, k^\prime \in \{1,...,K_T\}$.

In the few-domain generalization scenario, we have the access to a set of $M$ base tasks $B = \{T_m\}^{M}_{m=1}$, each sampled from task space $\mathcal{T}$ with known source \& target domains.
For the given novel task $T^*$, however, we only have access to the source domains $\mathcal{D}^{T^*}_{src} = \{D_k\}_{k=1}^{K}$ with limited heterogeneity.
That is, $K \ll K_{T_m}$ for $T_m \in B$.
Our goal is to learn a model $f: \mathcal{X} \to \mathcal{Y}$ on source domains of novel task that can generalize well to novel unseen target domains, with the knowledge and experience learned from base tasks.
Formally, the target problem can be defined as follows:
\begin{problem}[Few-domain generalization]
	We are given $M$ base tasks $B = \{T_m\}^{M}_{m=1}$ and $K$ source (training) domains $\mathcal{D}^{T^*}_{src} = \{D_k\}_{k=1}^{K}$ from novel task $T^*$, where the observed heterogeneity is limited for $T^*$ compared with that in base tasks.
	The goal of few-domain generalization is to learn a generalizable predictive model $f: \mathcal{X} \to \mathcal{Y}$ from the $K$ source domains to achieve a minimum out-of-distribution prediction error on \emph{unseen} target domains $\mathcal{D}_{tar}^{T^*}$: 
	\begin{equation}
		\min_{f} \, \mathbb{E}_{D \in \mathcal{D}_{tar}^{T^*}} \mathbb{E}_{(\mathbf{x},y) \in D} [\mathcal{L}(f(\mathbf{x}),y)],
	\end{equation}
	where $\mathbb{E}$ is the expectation and $\mathcal{L}: \mathcal{Y} \times \mathcal{Y} \to [0, \infty)$ is the loss function.
\end{problem}

For traditional \emph{homogeneous} DG problem, we usually assume a common label space.
However, in few-domain generalization, different tasks may have potentially disjoint label space, that is, $\exists k,k'\in[1,M]$ such that $\mathcal{Y}_k \neq \mathcal{Y}_{k'} \neq \mathcal{Y}_*$, where $\mathcal{Y}_*$ represents the label space of novel task.
Therefore, through the pre-training on base tasks, our principle target is to learn a robust representation $\theta^*(\cdot)$, which can be quickly adapted into the novel task through fine-tuning.

\subsection{A Simple Baseline: Episodic Training on Base Tasks}
Inspired by the few-shot learning, we introduce a MAML\cite{finn2017model}-based episodic training scheme to address the few-domain generalization problem.
Actually, in traditional domain generalization problem, there exist several attempts to leverage episodic training strategy to improve the OOD generalization performance by virtually splitting source domains into \emph{meta-train} and \emph{meta-test} domains, as depicted in previous works such as MLDG \cite{li2018learning}, MetaReg \cite{balaji2018metareg}, Feature-Critic \cite{li2019feature}, etc. 
However, as found in \cite{balaji2018metareg}, such training strategy is originally designed for fast task adaptation via the meta-learned weight initialization, which may not fit a zero-shot problem well like domain generalization.

For few-domain generalization with base tasks, a straight-forward approach is to apply such episodic training strategy onto all the base tasks successively.
Likewise, we can create \emph{meta-train} and \emph{meta-test} domains in accordance with the source and target domains of each base task, as shown in Figure \ref{fig:mldg_fdg}.
For simplicity, we leverage the training protocol of MLDG to illustrate this process.

\textbf{Meta-Train}: For a given base task $T_m$, the model is first updated on the \emph{meta-train} domains $\mathcal{D}_{src}^{T_m}$ with the loss function:
\begin{equation}
	\label{equ:mldg_inner}
	\mathcal{F}(\cdot)=\frac{1}{|\mathcal{D}_{src}^{T_m}|} \sum_{k=1}^{|\mathcal{D}_{src}^{T_m}|} \frac{1}{N_{k}} \sum_{j=1}^{N_{k}} \mathcal{L}_{\theta}\left(f(x_{j}^{(k)}), y_{j}^{(k)}\right).
\end{equation}
And we can get an intermediate parameter $\theta^\prime$ through a single gradient step $\theta^{\prime}=\theta-\alpha \nabla_{\theta}\mathcal{F}$.

\textbf{Meta-Test}: Then the model parameters are optimized for the performance of $\theta^{\prime}$ with respect to $\theta$ on the \emph{meta-test} domains $\mathcal{D}_{tar}^{T_m}$ with the loss function:
\begin{equation}
	\label{equ:mldg_outer}
	\mathcal{G}(\cdot)=\frac{1}{|\mathcal{D}_{tar}^{T_m}|} \sum_{k=1}^{|\mathcal{D}_{tar}^{T_m}|} \frac{1}{N_{k}} \sum_{j=1}^{N_{k}} \mathcal{L}_{\theta^{\prime}}\left(f(x_{j}^{(k)}), y_{j}^{(i)}\right).
\end{equation}

\textbf{Overall Objective}: Finally, the loss in \emph{meta-train} and \emph{meta-test} phases can be optimized simultaneously with the objective:
\begin{equation}
	\label{equ:mldg_all}
	\mathop{\arg\min}_{\theta} \mathcal{F}(\theta)+\beta \mathcal{G}\left(\theta-\alpha \nabla_{\theta}\mathcal{F}\right).
\end{equation}

With the meta pre-training on task $T_m$, the representation $\theta^*(\cdot)$ is supposed to be robust to domain shift characterized by its source and target domains.
Through iteratively sampling tasks from base tasks set $B$, the model is exposed to different shift patterns and should be more capable of generalizing to unseen target domains on novel tasks. 

\subsection{Episodic Training with Meta Adaptive Task Sampling}
By directly applying episodic training on base tasks, we assume that all the base tasks contribute equally to the generalization of novel task.
However, we argue that it seldom happens due to the very large feature space provided by deep neural networks, and therefore the semantic concepts of base tasks will inevitably influence the learning on novel tasks.
Intuitively, one can learn to recognize a wolf more quickly by pre-training on recognizing semantically similar concepts (e.g. dog) than other dissimilar ones, as noted by \cite{zhou2019Learning, zhou2020Learning} in the context of few-shot learning.

In contrast to their work, our goal is to characterize such semantic relationships at task level rather than class level.
For example, we want to differentiate a base task which performs mammal recognition from another one which performs vehicle recognition, given that the novel task is to classify several animals.
Specifically, for a given task $T_m$, we summarize its \emph{semantic concept} at domain-level by computing the average representation $\bar{z}_k$ of $D_k \in D^{T_m}$: $\bar{z}_k =\frac{1}{N_k}\sum_{i=1}^{N_k}\Theta_{\theta}(x^{(k)}_i)$
and aggregate the semantic concept over source domains $D^{T_m}_{src}$ to get the task-level semantic representation $\bar{z}^{T_m}_{src}=\frac{1}{|\mathcal{D}_{src}^{T_m}|}\sum_{k=1}^{|\mathcal{D}_{src}^{T_m}|}\bar{z}_k$.
We can therefore define the \emph{semantic similarity} between base task $T_m$ and novel task $T^*$ as:
\begin{equation}
	\label{equ:mean_define}
	s(m) = cosine(\bar{z}^{T_m}_{src}, \bar{z}^{T^*}_{src}).	
\end{equation}

In addition to the discovery of similar semantic space, we want to further encourage the exploitation of similar shift patterns to the novel task among base tasks.
Specifically, we can also define the \emph{domain-shift similarity} by the best match between concept shift in base task $T_m$ and the observed domain shift in the source domains of novel task $T^*$ as follows:
\begin{equation}
	\label{equ:shift_define}
	q(m) = \mathop{\max}_{k,k' \in [1, K]} cosine(\bar{z}_{k}-\bar{z}_{k'}, \bar{z}_{tar}^{T_m}-\bar{z}_{src}^{T_m}).	
\end{equation}

In summary, we define the overall similarity between base task $T_m$ and novel task $T^*$ as:
\begin{equation}
	sim(m) = s(m) + \gamma q(m),
\end{equation}
and accordingly propose a \textbf{Meta Adaptive Task Sampling} (MATS) procedure to sample base tasks from task pool $B$ with the probability $p(m)$ computed as its normalized similarity:
\begin{equation}
	\label{equ:MATS}
	p(m) = \frac{sim(m)}{\sum^{M}_{l=1}sim(l)}.
\end{equation}
We describe the pipeline of our full method in Algorithm \ref{alg:mats}.

\begin{small}
\begin{algorithm}[t]
	\caption{Episodic Training with Meta Adaptive Task Sampling}
	\label{alg:mats}
	\begin{algorithmic}[1]
		\State \textbf{Input:} Base tasks set $B = [T_1, T_2, \dots, T_M]$, source domains of novel task $D_{src}^{T^*} = [D_1, D_2, \dots, D_K]$ and hyperparameters $\alpha$, $\beta$, $\gamma$, $\eta$.
		\State \textbf{Initialize model parameters}: $\theta$, $\psi$, $\psi_1, \psi_2, \dots, \psi_M$, where $\psi$ represents classifier parameter for the novel task, $\psi_m$ represents specific classifier parameter corresponding to different base tasks.
		\State \textbf{Pre-training Phase}:
		\While{not done training} 
		\State Sample a base task $T_m$ from $B$ with probability distribution defined as Eq.\ref{equ:MATS}.
		\State Update $\theta := \theta - \eta \nabla_{\theta}(\mathcal{F}(\theta)+\beta \mathcal{G}\left(\theta-\alpha \nabla_{\theta}\mathcal{F}\right))$
		\State Update $\psi_m := \psi_m - \eta \nabla_{\psi_m}(\mathcal{F}(\psi_m)+\beta \mathcal{G}\left(\psi_m-\alpha \nabla_{\psi_m}\mathcal{F}\right))$
		\EndWhile
		\State \textbf{Fine-tuning Phase}:
		\While{not done training} 
		\State Sample a batch from novel training data $D_{src}^{T^*}$
		\State Update $\theta := \theta - \eta \nabla_{\theta}(\mathcal{F}(\theta))$
		\State Update $\psi := \psi - \eta \nabla_{\psi}(\mathcal{F}(\psi))$
		\EndWhile
		\State \textbf{Output:} $\theta^*$, $\psi^*$
	\end{algorithmic}
\end{algorithm}
\end{small}

\section{Experiments}
\label{sec:exp}

\begin{table*}[ht]
	\centering
	\resizebox{0.9\textwidth}{!}{
        \begin{tabular}{@{}c|ccc|ccc|ccc|ccc@{}}
        \toprule
         & \multicolumn{3}{c|}{PACS} & \multicolumn{3}{c|}{VLCS} & \multicolumn{3}{c|}{OfficeHome} & \multicolumn{3}{c}{DomainNet} \\ \midrule
        Methods     & $K=1$    & $K=2$   & $K=3$   & $K=1$    & $K=2$   & $K=3$   & $K=1$      & $K=2$     & $K=3$     & $K=1$     & $K=2$     & $K=3$    \\ \midrule
        ERM         & 49.0    & 67.2   & 71.8   & 59.6    & 67.6   & 72.1   & 49.2      & 54.2     & 55.5     & 73.3     & 82.4     & 85.3    \\
        JiGen       & 49.2    & 67.7   & 72.3   & 59.4    & 67.5   & 72.8   & 48.8      & 53.7     & 55.3     & 74.2     & 81.7     & 84.6    \\
        RSC         & 49.7    & 68.7   & 73.4   & 60.6    & 68.9   & 72.7   & 46.9      & 52.0     & 54.3     & 73.8     & 81.7     & 84.6    \\ \bottomrule
        \end{tabular}
	}
	\caption{OOD performance when varing the number of training domains $K$.}
	\label{table:stage1}
\end{table*}

The primary goal of our experimental evaluation is to answer the following questions:
\begin{itemize}
    \item Does the heterogeneity of training data (e.g. number of seen domains) influence the OOD generalization ability of traditional DG methods?
    \item Is the newly introduced few-domain generalization (FDG) framework beneficial when the heterogeneity of novel task is limited?
    \item Do different base tasks contribute equally to the novel task? If not, can we leverage it to improve the OOD generalization performance more efficiently?
\end{itemize}

\subsection{Experimental Settings}

\subsubsection{Datasets}
For our experiments, we consider several benchmark datasets widely used in domain generalization: 

\textbf{PACS}\cite{li2017deeper} consists of 4 domains: photo, art\_painting, cartoon and sketch. 
These domains depict the distribution shift induced by style transfer. 
It contains 9,991 examples of 7 classes including dog, elephant, giraffe, guitar, horse, house, person.

\textbf{VLCS}\cite{fang2013unbiased} aggregates photos from Caltech, LabelMe, Pascal VOC 2007 and SUN09. 
It formulates a classification task with 5 common classes: bird, car, chair, dog, person and contains 10,729 examples.

\textbf{Office-Home}\cite{venkateswara2017Deep} comprises 4 domains including art, clipart, product and real, with 65 classes. 
It contains 15,588 examples.

\textbf{DomainNet}\cite{peng2019moment} is currently the biggest public dataset for domain generalization. 
It contains 6 domains: clipart, infograph, painting, quickdraw, real and sketch. 
Similar to PACS, domain shift in it is characterized by style transfer. 
There are 586,575 samples and 345 classes in DomainNet.

\textbf{RMNIST}\cite{ghifary2015domain}(Rotated MNIST) is created from the well-known handwritten digits dataset MNIST by rotating certain degrees ranging from 0 to 75 every 15 degrees, thus generating 6 synthetic domains. 
It contains 70,000 examples.

%Considering the large class number and sample size of DomainNet and Office-Home, we randomly sample 7 classes from them respectively, to match the cardinality and difficulty of other datasets. 
For datasets with existing train-val-test splits generated by their maker, like PACS and VLCS, we follow their standard protocols. 
For other datasets, we set the split ratio as train: val: test = 8: 1: 1 and split them by ourselves.

\subsubsection{Baselines and implementation details}
We compare our method with several famous model-agnostic domain generalization algorithms. 
Our baselines are as follows:

\textbf{ERM}\cite{vapnik1998statistical} (Empirical Risk Minimization) simply aggregates data from all domains and minimizes the sum of sample errors, which shows comparable performance against other carefully designed DG algorithms in standard settings\cite{gulrajani2020search}.

\textbf{JiGen}\cite{carlucci2019domain} acts in a self-supervised manner, trying to solve the extra task of a jigsaw puzzle. 
It strengthens the spatial recognition ability of models by learning to restore from spatial permutations. 
It simultaneously optimizes training loss for object classification and permutation classification.

\textbf{RSC}\cite{2020Self} is one of the state-of-the-art DG algorithms on well-known benchmarks like PACS. 
Its training process bears resemblance to dropout, as it iteratively discards dominant features and force the other features to be activated. 

\textbf{MLDG}\cite{li2018learning} starts the fashion of applying meta-learning framework into domain generalization. 
We choose it as a baseline due to its simplicity and conformity of FDG problem.

As for implementation, we use resnet18\cite{he2016deep} as the backbone of above mentioned methods. 
We use SGD optimizer and set batch size to 32. 
For other hyperparameters, we follow the default settings used in the baselines' papers or codes. 

As mentioned in Sec \ref{subsec:setup}, different base tasks and the novel task may have disjoint label space, therefore the meta learning on base tasks actually acts as the pre-training, aiming at finding a robust and generalizable feature representation which can be quickly adapted to novel task with limited heterogeneity.
After pre-training, we simply finetune the whole network using the merged data from different domains of novel task, simulating the scenario where the heterogeneity of novel task is scarce and hard to perceive.

\subsection{Experimental Results}

\subsubsection{The influence of data heterogeneity on generalization performance}
To investigate the influence of data heterogeneity on model's OOD generalization ability, we evaluate several prevailing domain generalization methods on commonly used benchmarks. 
For the simplicity and fairness of comparison, we only use 4 domains (real, painting, clipart and sketch) of DomainNet in this experiment to match the cardinality and difficulty of other datasets. 
Likewise, we sample 7 classes from Office-Home and DomainNet respectively, to mitigate the effect of class number of datasets. 
Different from the standard evaluation process that uses "leave-one-domain-out" scenario, we vary the number of training domains $K$ from 1 to 3, simulating the different levels of heterogeneity, and calculate the average performance over the left domains.
We repeat every setting 5 times with different random seeds and calculate OOD accuracy for each training domain size $K$ by averaging all the possible split of training \& testing domains.

\begin{figure}[ht]
	\centering
	\begin{subfigure}{0.48\linewidth}
		\includegraphics[width=1\linewidth]{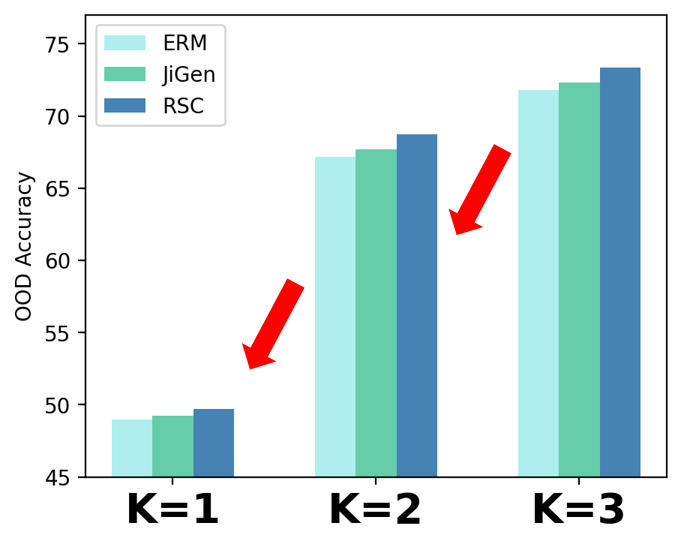}
		\caption{PACS}
		\label{fig:stage1pacs}
	\end{subfigure}
	\begin{subfigure}{0.48\linewidth}
		\includegraphics[width=1\linewidth]{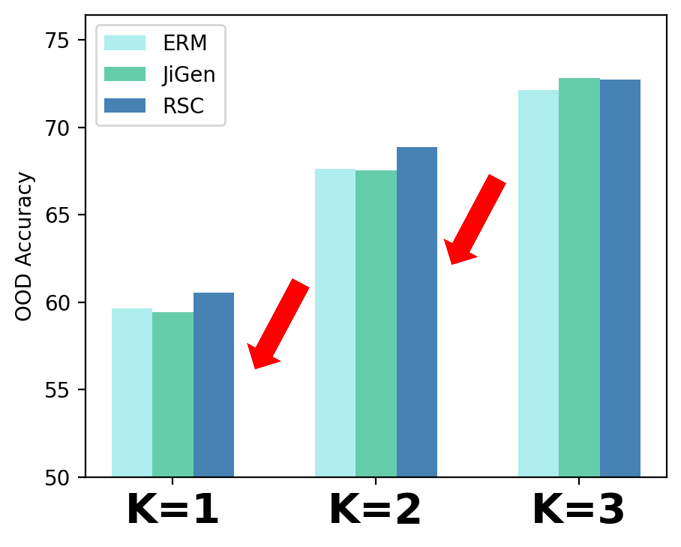}
		\caption{VLCS}
		\label{fig:stage1vlcs}
	\end{subfigure}
	\quad
	\begin{subfigure}{0.48\linewidth}
		\includegraphics[width=1\linewidth]{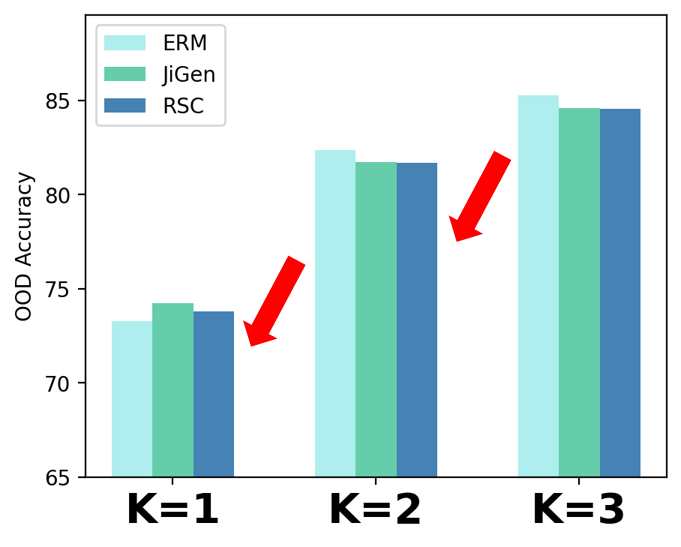}
		\caption{DomainNet}
		\label{fig:stage1domainnet}
	\end{subfigure}
	\begin{subfigure}{0.48\linewidth}
		\includegraphics[width=1\linewidth]{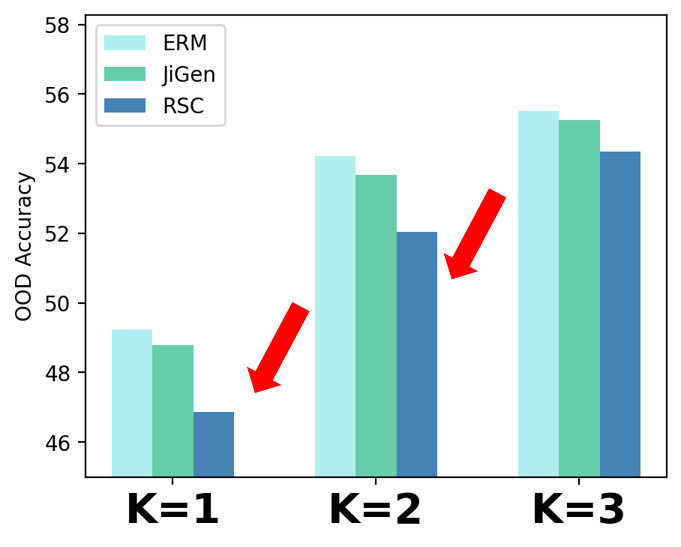}
		\caption{OfficeHome}
		\label{fig:stage1officehome}
	\end{subfigure}
	\caption{OOD performance when varing the number of training domains $K$.}
	\label{fig:stage1}
\end{figure}

From the experimental results shown in Table \ref{table:stage1} and Figure \ref{fig:stage1}, we have the following observations:
(1) For all the benchmarks and baselines, there exsit consistent trends that the OOD generalization performance deteriorates remarkably when the number of training domains $K$ decreases, proving the strong sensitivity of conventional methods to the data heterogeneity and coinciding with the theoretical analysis in \cite{2021How}.
(2) All the baselines perform comparably when the heterogeneity is sufficient (e.g. domain number $K=3$), which further verifies the claim in \cite{gulrajani2020search} that ERM could provide a competitive result when the training domains are diverse and all the methods are tuned carefully. 
(3) When the data heterogeneity is limited (e.g. $K=1$ as single DG setting), all the methods suffer from unstatisfactory results, which further reminds us the urgency of investigating suitable and practical method under few available domains.

\begin{table*}[ht]
	\centering
	\resizebox{0.9\textwidth}{!}{%
        \begin{tabular}{@{}c|ccc|ccc|ccc|ccc@{}}
        \toprule
        Methods    & \multicolumn{3}{c|}{ERM}                         & \multicolumn{3}{c|}{JiGen}                       & \multicolumn{3}{c|}{RSC}                          & \multicolumn{3}{c}{MLDG}                          \\ \midrule
        \#domains  & $K=1$           & $K=2$           & $K=3$           & $K=1$           & $K=2$           & $K=3$           & $K=1$            & $K=2$           & $K=3$           & $K=1$            & $K=2$           & $K=3$           \\ \midrule
        None       & 48.38          & 75.44          & 78.13          & 48.38          & 75.44          & 78.13          & 48.38           & 75.44          & 78.13          & 48.38           & 75.44          & 78.13          \\ \midrule
        VLCS       & +0.57          & +0.04          & -1.91          & -0.83          & -0.01          & -1.66          & +0.02           & +0.15          & -0.63          & +1.03           & +0.95          & -0.80          \\
        RMNIST     & +0.02          & -0.41          & -1.47          & -0.96          & -1.48          & -1.10          & +0.96           & -0.29          & -0.24          & +1.60           & +0.33          & -0.51          \\
        OfficeHome & +1.69          & -0.37          & +0.50          & +2.07          & -0.06          & +0.31          & +1.60           & -0.89          & -0.38          & +2.73           & -0.05          & +0.72          \\
        DomainNet  & \textbf{+9.52} & \textbf{+4.39} & \textbf{+2.85} & \textbf{+9.83} & \textbf{+4.46} & \textbf{+3.47} & \textbf{+14.20} & \textbf{+4.94} & \textbf{+3.06} & \textbf{+12.96} & \textbf{+5.14} & \textbf{+3.46} \\ \midrule
        Average    & +2.95          & +0.91          & -0.01          & +2.53          & +0.73          & +0.26          & +4.19           & +0.98          & +0.45          & \textbf{+4.58}  & \textbf{+1.59} & \textbf{+0.72} \\ \bottomrule
        \end{tabular}
	}
	\caption{Comparisons of different base tasks under FDG framework on PACS.}
	\label{table:stage2}
\end{table*}

\subsubsection{The effectiveness of FDG framework} 
In this experiment, we want to investigate whether the introduced FDG framework of leveraging base tasks as pre-training could improve the OOD generalization performance on novel tasks, and whether different base tasks contribute equally to novel tasks. 
Specifically, we construct the novel tasks from PACS and base tasks from remaining four datasets DomainNet, OfficeHome, VLCS and RMNIST.
We apply four baselines ERM, JiGen, RSC and MLDG as pre-training methods on each base task and keep the learned representation, then we finetune the whole network on the training domains of novel tasks with varing number of available domains $K={1,2,3}$. 
Finally, we test the model on the unseen target domains of novel tasks.
For ablation, we also apply baselines directly onto the novel tasks without pre-training on base tasks, following conventional DG settings.
We repeat every setting 3 times with different random seeds and calculate OOD accuracy for each training domain size $K$ by averaging all the possible split of training \& testing domains.

From the results in Table \ref{table:stage2}, we can find that:
(1) With the pre-training on base tasks, all the baselines show the substantial improvement in terms of OOD generalization performance on the novel tasks compared with those without pre-training (None), which prove the effectiveness of few-domain generalization framework.
That is to say, for domain generalization, or more generally OOD generalization, pre-training on datasets with heterogeneity can still be beneficial despite the fact that the model is pre-trained on very large natural datasets like ImageNet.
(2) As the number of source training domains $K$ decreases, the improvement brought by FDG framework clearly increases, indicating the more favourable nature of FDG framework in few-domain settings.
(3) Among all the baselines, MLDG perferms best, possibly because the explicit modeling of domain shift in meta-learning scheme.
Such training strategy enables the model to proactively resist domain shift.
(4) Though the involvement of base tasks generally enhances the model robustness, different base tasks contribute unequally to the novel task.
For example, pre-training on RMNIST and VLCS do not help much in terms of generalization on PACS, while pre-training on the base tasks from DomainNet improve the vanilla DG baselines by almost 20\% relatively. 
Such findings demonstrate that few-domain generalization can not be addressed by simply pre-training on all the available base tasks without differentiate them according to their relationships with novel task, leading to our proposed MATS.

\begin{table}[ht]
	\centering
	\resizebox{0.4\textwidth}{!}{%
        \begin{tabular}{@{}cc|ccccc@{}}
        \toprule
        source      & target     & ERM  & JiGen & RSC           & MLDG          & MATS          \\ \midrule
        P           & A+C+S      & 42.3 & 41.1  & \textbf{45.0} & 44.5          & \textbf{44.8} \\
        A           & P+C+S      & 67.6 & 66.7  & 67.9          & 68.3          & \textbf{68.5} \\
        C           & P+A+S      & 70.3 & 70.8  & 72.5          & 72.4          & \textbf{72.8} \\
        S           & P+A+C      & 37.0 & 34.4  & 37.2          & 40.4          & \textbf{40.8} \\ \midrule
        \multicolumn{2}{c|}{$K=1$} & 54.3 & 53.3  & 55.6          & 56.4          & \textbf{56.7} \\ \midrule
        P+A         & C+S        & 54.8 & 53.1  & 58.3          & 57.3          & \textbf{59.8} \\
        P+C         & A+S        & 72.4 & 70.8  & 72.9          & 73.8          & \textbf{74.9} \\
        P+S         & A+C        & 66.2 & 66.8  & 66.2          & 68.5          & \textbf{70.4} \\
        A+C         & P+S        & 82.0 & 82.5  & 83.4          & 83.5          & \textbf{84.6} \\
        A+S         & P+C        & 83.6 & 84.6  & 84.2          & 84.9          & \textbf{85.4} \\
        C+S         & P+A        & 76.8 & 77.2  & \textbf{78.2} & 76.6          & \textbf{77.7} \\ \midrule
        \multicolumn{2}{c|}{$K=2$} & 72.6 & 72.5  & 73.8          & 74.1          & \textbf{75.5} \\ \midrule
        P+A+C       & S          & 70.4 & 72.0  & 73.7          & 73.9          & \textbf{76.5} \\
        P+A+S       & C          & 73.9 & 74.8  & 74.0          & \textbf{76.1} & \textbf{76.4} \\
        P+C+S       & A          & 77.3 & 77.3  & 78.5          & 76.9          & \textbf{78.9} \\
        A+C+S       & P          & 95.0 & 95.2  & 95.2          & 95.3          & \textbf{95.7} \\ \midrule
        \multicolumn{2}{c|}{$K=3$} & 79.1 & 79.8  & 80.3          & 80.5          & \textbf{81.8} \\ \bottomrule
        \end{tabular}
	}
	\caption{Comparisons of different methods on PACS.}
	\label{table:stage3k1}
\end{table}

\begin{table}[ht]
	\centering
	\resizebox{0.4\textwidth}{!}{%
        \begin{tabular}{@{}cc|lllll@{}}
        \toprule
        source      & target     & \multicolumn{1}{c}{ERM} & \multicolumn{1}{c}{JiGen} & \multicolumn{1}{c}{RSC} & \multicolumn{1}{c}{MLDG} & \multicolumn{1}{c}{MATS} \\ \midrule
        S           & C+R+P      & 77.2                    & 78.5                      & 76.3                    & 79.7                     & \textbf{80.3}            \\
        C           & S+R+P      & 68.7                    & 68.6                      & 69.7                    & 70.8                     & \textbf{71.5}            \\
        R           & S+C+P      & 72.8                    & 71.5                      & 72.3                    & 72.9                     & \textbf{74.5}            \\
        P           & S+C+R      & 78.2                    & 77.6                      & 77.6                    & 78.9                     & \textbf{80.4}            \\ \midrule
        \multicolumn{2}{c|}{$K=1$} & 74.2                    & 74.0                      & 74.0                    & 75.6                     & \textbf{76.7}            \\ \midrule
        S+C         & R+P        & 82.9                    & 83.1                      & 83.3                    & 83.9                     & \textbf{85.2}            \\
        S+R         & C+P        & 81.0                    & 81.5                      & 82.3                    & 84.4                     & \textbf{85.2}            \\
        S+P         & C+R        & 88.2                    & 87.3                      & 88.8                    & 89.2                     & \textbf{91.8}            \\
        C+R         & S+P        & 78.0                    & 79.2                      & 79.4                    & 80.0                     & \textbf{82.8}            \\
        C+P         & S+R        & 87.2                    & 86.8                      & 86.9                    & 88.3                     & \textbf{88.8}            \\
        R+P         & S+C        & 74.9                    & 74.6                      & 75.1                    & 78.2                     & \textbf{79.9}            \\ \midrule
        \multicolumn{2}{c|}{$K=2$} & 82.0                    & 82.1                      & 82.6                    & 84.0                     & \textbf{85.6}            \\ \midrule
        S+C+R       & P          & 82.1                    & 82.6                      & 82.1                    & 84.0                     & \textbf{85.7}            \\
        S+C+P       & R          & 93.8                    & 93.3                      & 93.9                    & \textbf{94.2}            & 93.8                     \\
        S+R+P       & C          & 86.1                    & 84.0                      & 83.7                    & 86.7                     & \textbf{89.6}            \\
        C+R+P       & S          & 83.1                    & 81.1                      & 82.1                    & 83.1                     & 83.8                     \\ \midrule
        \multicolumn{2}{c|}{$K=3$} & 86.3                    & 85.2                      & 85.5                    & 87.0                     & \textbf{88.2}            \\ \bottomrule
        \end{tabular}
	}
	\caption{Comparisons of different methods on DomainNet.}
	\label{table:stage3k2}
\end{table}

\begin{figure}[ht]
	\centering
	\begin{subfigure}{0.49\linewidth}
		\includegraphics[width=1\linewidth]{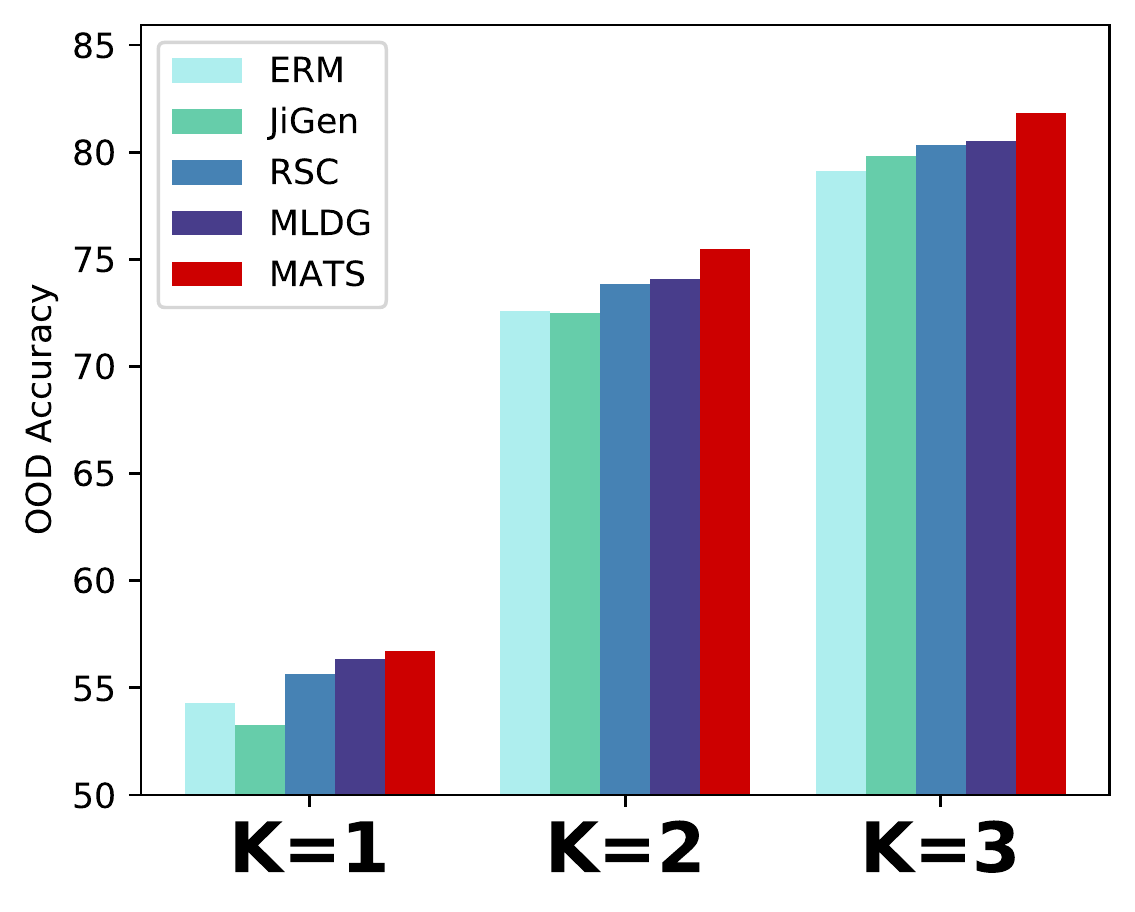}
		\caption{PACS}
		\label{fig:stage3pacs}
	\end{subfigure}
	\begin{subfigure}{0.49\linewidth}
		\includegraphics[width=1\linewidth]{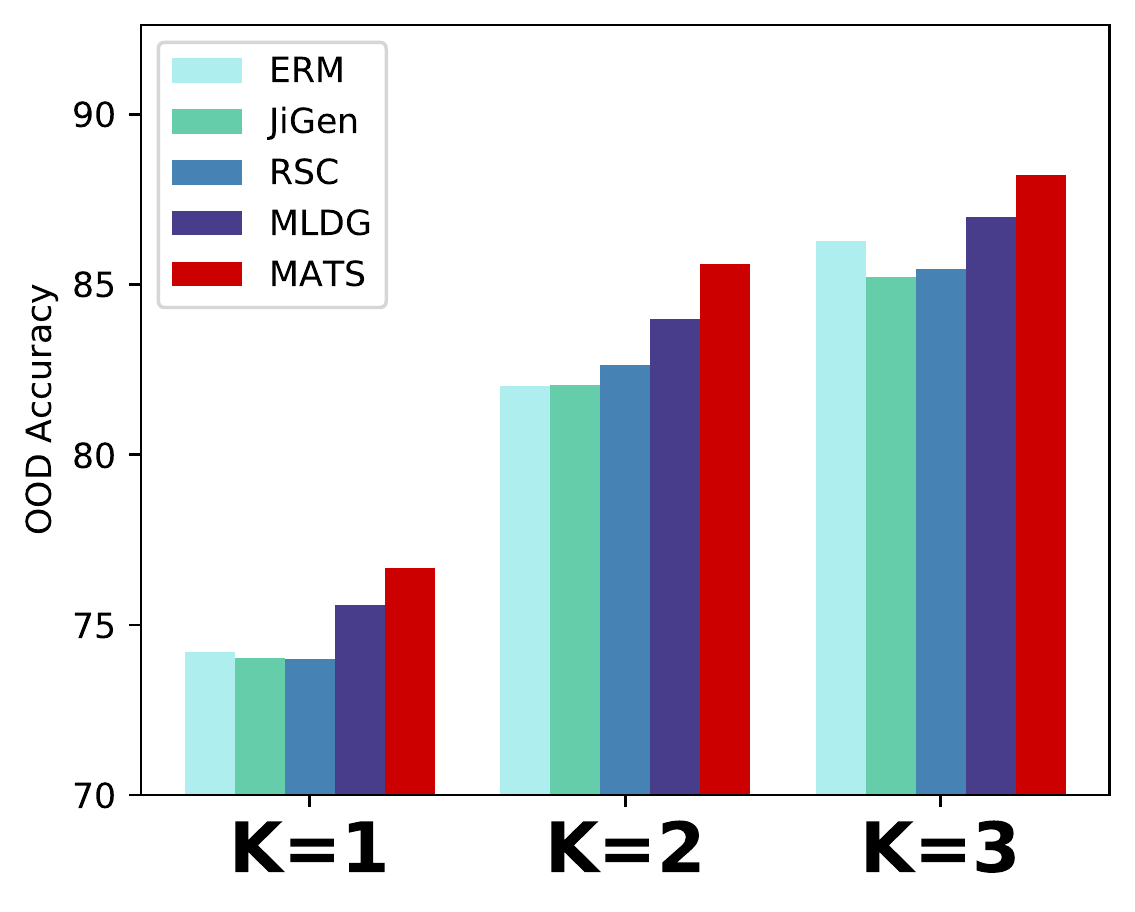}
		\caption{DomainNet}
		\label{fig:stage3domainnet}
	\end{subfigure}
	\caption{Comparisions of different methods on PACS and DomainNet.}
	\label{fig:stage3}
\end{figure}

\begin{table*}[ht]
	\centering
	\resizebox{0.8\textwidth}{!}{
    \begin{tabular}{@{}c|c|ccccccccccc@{}}
    \toprule
                     & Data & 5\%           & 10\%          & 20\%          & 30\%          & 40\%          & 50\%          & 60\%          & 70\%          & 80\%          & 90\%          & 100\%         \\ \midrule
    \multirow{3}{*}{$K=1$} & ERM  & 30.3          & 33.3          & 48.7          & 52.0          & 52.7          & 52.8          & 54.1          & 52.4          & 52.8          & 54.2          & 54.3          \\
                     & MLDG & 33.3          & 35.3          & 52.8          & 54.7          & 54.9          & 56.0          & 55.7          & 55.4          & 55.6          & 57.1          & 56.4          \\
                     & MATS & \textbf{33.4} & \textbf{35.7} & \textbf{53.4} & \textbf{54.7} & \textbf{55.1} & \textbf{56.2} & \textbf{56.6} & \textbf{55.6} & \textbf{56.8} & \textbf{57.4} & \textbf{56.7} \\ \midrule
    \multirow{3}{*}{$K=2$} & ERM  & 55.9          & 59.0          & 67.8          & 69.6          & 71.1          & 72.4          & 71.9          & 71.8          & 71.8          & 72.3          & 72.6          \\
                     & MLDG & 59.5          & 62.8          & 70.0          & 71.4          & 73.1          & 73.3          & 74.7          & 74.1          & 74.4          & 74.5          & 74.1          \\
                     & MATS & \textbf{61.9} & \textbf{65.1} & \textbf{71.8} & \textbf{72.7} & \textbf{73.8} & \textbf{75.3} & \textbf{75.9} & \textbf{75.0} & \textbf{75.4} & \textbf{75.4} & \textbf{75.5} \\ \midrule
    \multirow{3}{*}{$K=3$} & ERM  & 63.1          & 66.0          & 74.3          & 75.7          & 77.4          & 78.4          & 78.5          & 77.3          & 78.4          & 78.3          & 79.1          \\
                     & MLDG & 66.8          & 69.8          & 76.9          & 77.5          & 79.0          & 79.7          & 79.9          & 79.8          & 80.0          & 80.2          & 80.5          \\
                     & MATS & \textbf{69.1} & \textbf{72.1} & \textbf{77.9} & \textbf{78.7} & \textbf{80.0} & \textbf{80.8} & \textbf{81.1} & \textbf{80.7} & \textbf{80.7} & \textbf{80.8} & \textbf{81.8} \\ \bottomrule
    \end{tabular}}
	\caption{Comparisons between ERM, MLDG and MATS when available data of novel task change}
	\label{table:consistency}
\end{table*}

\subsubsection{The effectiveness of MATS}
In this experiment, we want to validate the performance of different methods given all the base tasks.
Specifically, we construct novel tasks from PACS (or DomainNet) and base tasks from the other left datasets.
By merging all the base tasks into a task pool, we apply different methods as pre-training and keep the learned representation space.
We then finetune the whole model on the source domains of novel tasks and test it on the unseen target domains.
Every experiment is repeated 3 times with different random seeds. 

From the results in Table \ref{table:stage3k1}, Table \ref{table:stage3k2} and Figure \ref{fig:stage3}, we can find that:
(1) Our proposed algorithm MATS generally surpasses all the baselines on most of the settings, proving the effectiveness of base task choice strategy.
That is to say, combining the episodic training stategy with task selection mechanism, we can exploit the historical experiences more efficiently to achieve OOD generalization on new tasks.
(2) MATS shows the most clear margin on when $K=2$, which verifies the fact that it is suitable for the scenarios when the perceived heterogeneity is limited.
(3) MATS can outperform meta-learning baseline MLDG even when there is no obeserved heterogeneity ($K=1$), which demonstrate the effecacy of \emph{semantic similarity} in task selection process.

\subsection{Ablation Study}

\subsubsection{Availability of data}
In this experiment, we simulate a scenario where there are only a part of data available and further investigate the performance consistency of our proposed MATS.
Specifically, we vary the available proportion of data from 5\% to 100\%, test the proposed MATS along with the MLDG and ERM. 

From the results in Table \ref{table:consistency}, MATS consistently outperforms baselines over different amount of available data, which demonstrates its potential on some data costly applications like healthcare.

\subsubsection{Performance with/without domain-shift similarity}
In this experiment, we want to validate the effectiveness of the proposed \emph{domain-shift similarity} in Equ \ref{equ:shift_define}.
Specifically, we set $\gamma=0$ as an ablation where we only use \emph{semantic similarity}.

From the results in Table \ref{table:ablation}, we can find the MATS without \emph{domain-shift similarity} generally outperforms the MLDG and the full MATS further improves the previous one, which validates the effectiveness of both two similarity measures.

\begin{table}[ht]
	\centering
	\resizebox{0.45\textwidth}{!}{%
        \begin{tabular}{@{}c|c|ccc@{}}
        \toprule
                                   & Methods & MLDG & MATS ($\gamma=0$) & MATS \\ \midrule
        \multirow{3}{*}{PACS}      & $K=1$    & 56.4 & 56.7     & \textbf{56.7} \\
                                   & $K=2$    & 74.1 & 74.8     & \textbf{75.5} \\
                                   & $K=3$    & 80.5 & 81.2     & \textbf{81.8} \\ \midrule
        \multirow{3}{*}{DomainNet} & $K=1$    & 75.6 & 76.7     & \textbf{76.7} \\
                                   & $K=2$    & 84.0 & 84.7     & \textbf{85.6} \\
                                   & $K=3$    & 87.0 & 87.4     & \textbf{88.2} \\ \bottomrule
        \end{tabular}
	}
	\caption{Ablation study for similarity metrics.}
	\label{table:ablation}
\end{table}

\section{Conclusion}
\label{sec:conclusion}
We propose \emph{few-domain generalizaion}, a framework which aims for OOD generalization when there is limited  data heterogeneity, leveraging previous tasks.
We prove that this framework boosts the OOD generalization performance on novel tasks. 
Considering the fact that different base tasks contribute unequally to a specific novel task, current methods which do not differentiate base tasks may result in low efficiency. 
To address this issue, We furtherly propose Meta Adaptive Task Sampling (MATS) procedure, which takes into account of the semantic and domain shift similarity between base tasks and the novel task.
We demonstrate the effectiveness of MATS on benchmarks including 
PACS and DomainNet, where MATS outperforms several state-of-the-art DG baselines.

%%%%%%%%% REFERENCES
{\small
\bibliographystyle{ieee_fullname}
\bibliography{egbib}
}

\section{Appendix}

\label{sec:appendix}
% In appendix, we provide more empirical results that could support our claims at the beginning of Sec.4.

\subsection{Influence of different base tasks on novel task PACS}
In this experiment, we investigate how does choice of base task influence the performance on novel task.
We sample two 7-classes subsets of OfficeHome and DomainNet respectively.
For DomainNet, we enumerate several 4-domains subsets among all the 6 available domains to further investigate the influence of domain similarity between base and novel task.
 
\begin{table*}[ht]
\centering
\caption{Contributions from different base tasks to novel tasks of PACS under FDG framework: ERM}
\label{table:stage2-ERM}
\resizebox{\textwidth}{!}{%
\begin{tabular}{@{}c|cccccccccccccc@{}}
\toprule
source            & P     & A     & C     & S     & P+A  & P+C  & P+S  & A+C  & A+S  & C+S  & P+A+C & P+A+S & P+C+S & A+C+S \\
target            & A+C+S & P+C+S & P+A+S & P+A+C & C+S  & A+S  & A+C  & P+S  & P+C  & P+A  & S     & C     & A     & P     \\ \midrule
None              & 35.5  & 62.4  & 66.4  & 29.4  & 47.3 & 66.4 & 62.4 & 80.9 & 83.4 & 75.1 & 68.8  & 72.2  & 76.0  & 95.6  \\ \midrule
VLCS              & 38.6  & 59.4  & 69.3  & 28.5  & 44.7 & 67.6 & 64.8 & 78.8 & 83.1 & 75.3 & 62.4  & 71.5  & 76.0  & 95.0  \\
RMNIST            & 33.7  & 63.4  & 66.5  & 30.0  & 48.0 & 66.6 & 61.9 & 80.2 & 83.0 & 75.1 & 65.1  & 72.4  & 74.7  & 94.4  \\
OfficeHome-v1     & 35.6  & 65.0  & 67.7  & 30.8  & 52.9 & 68.9 & 61.9 & 81.7 & 83.8 & 75.0 & 68.8  & 72.9  & 76.1  & 94.8  \\
OfficeHome-       & 39.4  & 62.9  & 68.9  & 29.1  & 49.5 & 70.5 & 62.0 & 81.2 & 83.2 & 73.8 & 71.3  & 71.8  & 76.4  & 95.0  \\
DomainNet-v1-qisc & 42.3  & 69.6  & 74.8  & 43.5  & 59.7 & 73.8 & 70.7 & 83.8 & 85.3 & 80.0 & 71.5  & 74.8  & 78.9  & 95.2  \\
DomainNet-v1-qpsc & 47.3  & 71.0  & 74.3  & 37.5  & 63.3 & 73.4 & 71.9 & 83.7 & 85.3 & 78.2 & 74.0  & 75.2  & 79.5  & 95.2  \\
DomainNet-v1-rpsc & 52.1  & 74.2  & 73.2  & 35.2  & 68.4 & 75.3 & 71.1 & 84.9 & 85.8 & 77.2 & 75.8  & 77.0  & 79.2  & 95.5  \\
DomainNet-v2-qisc & 37.1  & 61.5  & 65.3  & 29.3  & 48.7 & 70.0 & 59.0 & 80.4 & 82.6 & 71.7 & 68.3  & 71.9  & 76.3  & 93.8  \\
DomainNet-v2-qpsc & 37.5  & 62.7  & 66.9  & 31.3  & 52.7 & 69.9 & 60.5 & 81.3 & 83.1 & 75.1 & 69.0  & 73.4  & 77.4  & 94.1  \\
DomainNet-v2-rpsc & 35.4  & 67.8  & 66.9  & 35.7  & 57.6 & 68.9 & 59.7 & 80.6 & 82.2 & 74.3 & 68.7  & 72.4  & 76.5  & 93.8  \\ \midrule
average           & 39.9  & 65.8  & 69.4  & 33.1  & 54.5 & 70.5 & 64.4 & 81.7 & 83.7 & 75.6 & 69.5  & 73.3  & 77.1  & 94.7  \\ \bottomrule
\end{tabular}
}
\end{table*}

\begin{table*}[ht]
\centering
\caption{Contributions from different base tasks to novel tasks of PACS under FDG framework: MLDG}
\label{table:stage2-MLDG}
\resizebox{\textwidth}{!}{%
\begin{tabular}{@{}c|cccccccccccccc@{}}
\toprule
source            & P     & A     & C     & S     & P+A  & P+C  & P+S  & A+C  & A+S  & C+S  & P+A+C & P+A+S & P+C+S & A+C+S \\
target            & A+C+S & P+C+S & P+A+S & P+A+C & C+S  & A+S  & A+C  & P+S  & P+C  & P+A  & S     & C     & A     & P     \\ \midrule
None              & 35.5  & 62.4  & 66.4  & 29.4  & 47.3 & 66.4 & 62.4 & 80.9 & 83.4 & 75.1 & 68.8  & 72.2  & 76.0  & 95.6  \\ \midrule
VLCS              & 36.5  & 61.5  & 69.1  & 30.6  & 46.1 & 67.9 & 65.3 & 81.3 & 83.7 & 75.2 & 64.9  & 72.3  & 76.4  & 95.8  \\
RMNIST            & 37.2  & 62.3  & 67.4  & 33.1  & 48.0 & 70.0 & 64.2 & 81.9 & 82.9 & 74.2 & 68.8  & 72.7  & 74.7  & 94.3  \\
OfficeHome-v1     & 37.4  & 64.9  & 69.4  & 30.0  & 56.5 & 68.7 & 65.0 & 81.3 & 84.3 & 75.0 & 65.8  & 74.1  & 75.6  & 95.0  \\
OfficeHome-v2     & 41.0  & 67.1  & 67.6  & 28.7  & 55.6 & 70.1 & 64.1 & 81.7 & 82.7 & 73.1 & 71.7  & 72.4  & 76.7  & 94.6  \\
DomainNet-v1-qisc & 47.8  & 72.1  & 74.2  & 41.1  & 62.3 & 74.9 & 71.8 & 83.7 & 86.2 & 77.8 & 75.7  & 77.9  & 76.7  & 95.2  \\
DomainNet-v1-qpsc & 54.9  & 76.8  & 75.7  & 43.9  & 69.0 & 75.9 & 73.8 & 85.2 & 86.2 & 78.5 & 75.1  & 78.7  & 78.5  & 94.9  \\
DomainNet-v1-rpsc & 53.3  & 76.4  & 74.9  & 45.0  & 67.8 & 73.7 & 73.7 & 83.6 & 87.2 & 79.2 & 73.2  & 79.0  & 78.8  & 95.5  \\
DomainNet-v2-qisc & 39.2  & 67.2  & 68.6  & 32.3  & 56.0 & 71.4 & 64.3 & 81.9 & 83.3 & 74.2 & 73.0  & 72.9  & 76.7  & 93.9  \\
DomainNet-v2-qpsc & 37.8  & 68.2  & 70.2  & 33.2  & 55.8 & 71.8 & 63.6 & 82.5 & 83.2 & 74.9 & 72.3  & 71.9  & 76.8  & 94.6  \\
DomainNet-v2-rpsc & 39.0  & 66.5  & 69.9  & 30.4  & 54.8 & 70.8 & 65.7 & 82.6 & 83.9 & 75.4 & 71.7  & 73.5  & 76.4  & 95.0  \\ \midrule
average           & 42.4  & 68.3  & 70.7  & 34.8  & 57.2 & 71.5 & 67.1 & 82.6 & 84.3 & 75.8 & 71.2  & 74.5  & 76.7  & 94.9  \\ \bottomrule
\end{tabular}
}
\end{table*}

\begin{table*}[ht]
\centering
\caption{Contributions from different base tasks to novel tasks of PACS under FDG framework: JiGen}
\label{table:stage2-JiGen}
\resizebox{\textwidth}{!}{%
\begin{tabular}{@{}c|cccccccccccccc@{}}
\toprule
source            & P     & A     & C     & S     & P+A  & P+C  & P+S  & A+C  & A+S  & C+S  & P+A+C & P+A+S & P+C+S & A+C+S \\
target            & A+C+S & P+C+S & P+A+S & P+A+C & C+S  & A+S  & A+C  & P+S  & P+C  & P+A  & S     & C     & A     & P     \\ \midrule
None              & 35.5  & 62.4  & 66.4  & 29.4  & 47.3 & 66.4 & 62.4 & 80.9 & 83.4 & 75.1 & 68.8  & 72.2  & 76.0  & 95.6  \\ \midrule
VLCS              & 37.3  & 56.3  & 69.0  & 27.5  & 44.4 & 67.6 & 64.0 & 79.3 & 83.3 & 75.1 & 63.3  & 72.2  & 75.2  & 95.2  \\
RMNIST            & 34.6  & 63.4  & 66.1  & 25.6  & 48.2 & 65.5 & 59.8 & 79.7 & 82.4 & 73.9 & 67.1  & 71.8  & 74.9  & 94.4  \\
OfficeHome-v1     & 36.7  & 64.1  & 69.1  & 33.9  & 53.2 & 69.5 & 62.4 & 82.2 & 83.6 & 74.3 & 69.7  & 73.4  & 76.0  & 95.2  \\
OfficeHome-v2     & 39.3  & 62.3  & 68.7  & 31.5  & 50.9 & 69.5 & 62.7 & 81.9 & 83.0 & 73.9 & 70.7  & 72.4  & 75.7  & 95.0  \\
DomainNet-v1-qisc & 42.3  & 69.9  & 74.5  & 43.3  & 60.7 & 73.4 & 70.6 & 83.9 & 85.4 & 78.5 & 74.1  & 76.5  & 79.7  & 94.8  \\
DomainNet-v1-qpsc & 48.1  & 71.8  & 74.0  & 41.5  & 65.0 & 75.2 & 72.5 & 84.2 & 85.3 & 78.4 & 74.6  & 77.1  & 79.3  & 95.8  \\
DomainNet-v1-rpsc & 53.9  & 72.1  & 73.8  & 33.5  & 67.9 & 74.9 & 70.9 & 85.2 & 85.9 & 77.9 & 76.4  & 76.7  & 78.7  & 95.7  \\
DomainNet-v2-qisc & 37.0  & 58.7  & 65.4  & 30.3  & 48.7 & 70.1 & 59.6 & 81.1 & 82.4 & 72.1 & 70.2  & 71.2  & 77.4  & 93.6  \\
DomainNet-v2-qpsc & 37.5  & 60.6  & 67.9  & 31.2  & 50.8 & 70.4 & 60.6 & 81.4 & 83.1 & 75.0 & 70.0  & 72.1  & 77.3  & 94.4  \\
DomainNet-v2-rpsc & 36.0  & 66.9  & 66.4  & 32.2  & 56.8 & 69.5 & 60.6 & 81.0 & 82.4 & 73.9 & 69.2  & 72.6  & 76.4  & 93.6  \\ \midrule
average           & 40.2  & 64.6  & 69.5  & 33.1  & 54.7 & 70.6 & 64.4 & 82.0 & 83.7 & 75.3 & 70.5  & 73.6  & 77.0  & 94.8  \\ \bottomrule
\end{tabular}
}
\end{table*}

\begin{table*}[h]
\centering
\caption{Contributions from different base tasks to novel tasks of PACS under FDG framework: RSC}
\label{table:stage2-RSC}
\resizebox{\textwidth}{!}{%
\begin{tabular}{@{}c|cccccccccccccc@{}}
\toprule
source            & P     & A     & C     & S     & P+A  & P+C  & P+S  & A+C  & A+S  & C+S  & P+A+C & P+A+S & P+C+S & A+C+S \\
target            & A+C+S & P+C+S & P+A+S & P+A+C & C+S  & A+S  & A+C  & P+S  & P+C  & P+A  & S     & C     & A     & P     \\ \midrule
None              & 35.5  & 62.4  & 66.4  & 29.4  & 47.3 & 66.4 & 62.4 & 80.9 & 83.4 & 75.1 & 68.8  & 72.2  & 76.0  & 95.6  \\ \midrule
VLCS              & 32.3  & 57.3  & 69.8  & 34.3  & 44.1 & 67.8 & 63.3 & 79.9 & 83.1 & 76.0 & 68.0  & 71.0  & 75.9  & 95.2  \\
RMNIST            & 35.7  & 64.5  & 68.0  & 29.2  & 51.1 & 68.8 & 62.6 & 82.1 & 83.1 & 72.8 & 70.1  & 73.8  & 73.9  & 93.7  \\
OfficeHome-v1     & 38.2  & 66.4  & 69.3  & 30.8  & 53.5 & 69.1 & 62.9 & 81.7 & 83.3 & 74.7 & 68.3  & 72.9  & 75.7  & 95.1  \\
OfficeHome-v2     & 37.0  & 64.6  & 68.2  & 30.2  & 55.0 & 70.4 & 60.6 & 81.7 & 82.3 & 73.6 & 68.6  & 71.5  & 76.3  & 94.7  \\
DomainNet-v1-qisc & 51.1  & 72.5  & 73.7  & 41.3  & 64.4 & 73.8 & 72.6 & 83.2 & 85.7 & 78.8 & 73.4  & 77.6  & 78.4  & 94.6  \\
DomainNet-v1-qpsc & 55.2  & 76.1  & 75.4  & 45.8  & 70.0 & 74.3 & 73.5 & 83.4 & 85.5 & 79.6 & 72.2  & 77.5  & 79.8  & 94.3  \\
DomainNet-v1-rpsc & 59.8  & 78.5  & 75.5  & 46.0  & 73.2 & 73.6 & 72.2 & 83.5 & 86.7 & 80.0 & 73.7  & 77.0  & 80.3  & 95.5  \\
DomainNet-v2-qisc & 36.4  & 62.8  & 65.1  & 26.8  & 49.5 & 67.9 & 59.4 & 80.9 & 81.9 & 69.8 & 70.3  & 71.8  & 75.1  & 93.2  \\
DomainNet-v2-qpsc & 35.3  & 60.8  & 68.1  & 23.5  & 48.2 & 69.0 & 62.1 & 80.2 & 81.5 & 73.5 & 71.0  & 72.1  & 75.5  & 92.9  \\
DomainNet-v2-rpsc & 39.1  & 65.5  & 64.8  & 31.1  & 58.0 & 67.0 & 59.6 & 78.7 & 81.3 & 71.7 & 67.6  & 71.6  & 75.5  & 93.4  \\ \midrule
average           & 42.0  & 66.9  & 69.8  & 33.9  & 56.7 & 70.2 & 64.9 & 81.5 & 83.4 & 75.0 & 70.3  & 73.7  & 76.6  & 94.2  \\ \bottomrule
\end{tabular}
}
\end{table*}

\end{document}